\documentclass[preprint,12pt,authoryear]{elsarticle}

\usepackage{amssymb}

\usepackage{amsmath}

\usepackage{microtype}
\usepackage{graphicx}
\usepackage{subcaption}
\usepackage{booktabs} 
\usepackage{hyperref}

\usepackage{mathtools}
\usepackage{amsthm}
\usepackage{multirow}

\usepackage{algorithm}
\usepackage{algorithmic}

\usepackage{booktabs}
\usepackage{placeins}   
\usepackage{caption}
\usepackage{array}
\usepackage{chngcntr}

\journal{CSDA}

\begin{document}

\begin{frontmatter}

\title{A Data-Informed Variational Clustering Framework for Noisy High-Dimensional Data}

\author[aff1]{Wan Ping Chen\corref{cor1}}
\cortext[cor1]{Corresponding author.}
\ead{159986@mail.fju.edu.tw}

\affiliation{organization={Department of Mathematics, Fu Jen Catholic University, Taipei, Taiwan},
            addressline={No. 510 Zhongzheng Rd., Xinzhuang Dist.}, 
            city={New Taipei City},
            postcode={242062}, 
            country={Taiwan}}

\begin{abstract}
Clustering in high-dimensional settings with severe feature noise remains challenging, especially when only a small subset of dimensions is informative and the final number of clusters is not specified in advance. In such regimes, partition recovery, feature relevance learning, and structural adaptation are tightly coupled, and standard likelihood-based methods can become unstable or overly sensitive to noisy dimensions. We propose DIVI, a data-informed variational clustering framework that combines global feature gating with split-based adaptive structure growth. DIVI uses informative prior initialization to stabilize optimization, learns feature relevance in a differentiable manner, and expands model complexity only when local diagnostics indicate underfit. Beyond clustering performance, we also examine runtime scalability and parameter sensitivity in order to clarify the computational and practical behavior of the framework. Empirically, we find that DIVI performs competitively under severe feature noise, remains computationally feasible, and yields interpretable feature-gating behavior, while also exhibiting conservative growth and identifiable failure regimes in challenging settings. Overall, DIVI is best viewed as a practical variational clustering framework for noisy high-dimensional data rather than as a fully Bayesian generative solution.
\end{abstract}

\begin{keyword}
high-dimensional clustering \sep feature selection \sep variational inference \sep model-based clustering \sep noisy data
\end{keyword}



\end{frontmatter}


\section{Introduction}

Clustering in high-dimensional data remains challenging in machine learning and statistics, with applications ranging from spectral analysis and bioinformatics to semantic representation learning. In many such settings, although the ambient dimension may be in the hundreds or thousands, the latent group structure is supported by only a subset of informative features. As a result, clustering becomes inherently coupled with feature relevance learning: a method must simultaneously recover a meaningful partition of the observations and identify which dimensions carry discriminative signal. As dimensionality grows relative to sample size, this joint problem becomes increasingly difficult, since noisy or high-variance features can flatten the objective landscape, destabilize optimization, and obscure the true cluster structure \citep{bellman1957dynamic, aggarwal2001surprising}.

Classical approaches such as K-means and Gaussian Mixture Models (GMMs) typically treat all dimensions as equally relevant to the clustering objective. This assumption becomes problematic in noisy high-dimensional regimes. From an identifiability perspective, overwhelming noise can make it difficult to distinguish genuine cluster structure from random variation \citep{pmlr-v38-azizyan15}. From an optimization perspective, the resulting objective may contain poor local optima or flat regions in which fitting the noise is nearly as attractive as fitting the underlying structure \citep{bishop2006prml, watanabe2009algebraic}. These issues are particularly severe in small-sample settings, where limited observations must support both structure discovery and noise suppression.

A natural response is to incorporate feature selection or dimensionality reduction. However, many existing pipelines separate feature selection from clustering, rely on post-hoc screening, or impose fixed structural assumptions that do not reflect the interaction between feature relevance and cluster formation \citep{dy2004feature, alelyani2014}. Bayesian nonparametric models provide flexibility by allowing the number of clusters to vary \citep{rasmussen1999infinite}, but in finite and noisy regimes they may still suffer from weak identifiability and substantial computational cost \citep{miller2014inconsistency}. Recent neural and differentiable clustering methods improve scalability, yet many remain sensitive to high-dimensional noise and do not explicitly incorporate feature-level inductive bias into the early stages of optimization \citep{saha2023end, pakman2020neural}.

Motivated by these challenges, we study high-dimensional clustering through the interaction between optimization dynamics, feature relevance, and structural complexity. In difficult noise regimes, uninformed training objectives can drift into broad uncertainty regions in which both structure growth and feature gating remain indecisive, leading to unstable partitions and diffuse relevance estimates. To address this, we propose DIVI, a data-informed variational clustering framework that combines three components: data-informed prior initialization, differentiable feature gating, and split-based adaptive structure growth. DIVI does not treat feature relevance as a disconnected preprocessing step; instead, it learns feature gating within the clustering objective while allowing model structure to expand only when local diagnostics indicate underfit. In this sense, DIVI is designed as an optimization-aware clustering framework for noisy high-dimensional data.

Beyond clustering performance, we also study runtime scalability and parameter sensitivity in order to clarify how feature regularization, split frequency, and optimization settings affect the empirical behavior of DIVI. These analyses are useful for understanding not only when the method performs well, but also when it becomes conservative or enters failure regimes under extreme dimensionality.

Taken together, the present study makes three main points.

\begin{itemize}
    \item \textbf{A practical variational formulation for noisy high-dimensional clustering.}
    We develop DIVI as a clustering framework that combines data-informed initialization, differentiable feature gating, and split-based adaptive structure growth, thereby linking partition recovery and feature relevance learning within a single optimization procedure.

    \item \textbf{An interpretable view of structure growth and feature regularization.}
    Rather than treating tuning parameters as opaque hyperparameters, we show that the main controls of DIVI play distinct empirical roles: split frequency primarily governs structural expansion, whereas KL scaling primarily governs feature parsimony. This perspective helps clarify both the practical behavior and the computational cost of the method.

    \item \textbf{Empirical evidence on matched, misspecified, and real high-dimensional settings.}
    Across synthetic stress tests, misspecified robustness experiments, and real-data benchmarks, DIVI exhibits competitive clustering performance in several noisy high-dimensional settings together with interpretable feature weighting, while also revealing identifiable limitation regimes that help delineate its practical scope.
\end{itemize}

\section{Method: DIVI}
\label{sec:method}

DIVI is designed for high-dimensional clustering settings in which optimization stability, feature relevance, and structural adaptation must be handled jointly under severe feature noise. Rather than treating feature selection as a disconnected preprocessing step, DIVI combines global feature gating with point-estimated mixture components in a split-based variational clustering framework. Cluster assignments are handled implicitly through the mixture likelihood, while model structure is allowed to expand adaptively during training. In this sense, the current formulation is best understood as a practical, optimization-aware clustering framework with point-estimated mixture components and variationally treated feature gates.

\subsection{Feature-Gated Mixture Model}

Let $X=\{x_i\}_{i=1}^N$ with $x_i \in \mathbb{R}^D$. In high-dimensional settings, only a subset of coordinates may be informative for cluster separation, while the remaining coordinates behave as nuisance variation. To model this distinction, we introduce a global feature-relevance indicator $\phi_j \in \{0,1\}$ for each dimension $j=1,\dots,D$. When $\phi_j=1$, feature $j$ is modeled by a cluster-specific Gaussian component; when $\phi_j=0$, it is explained by a global background distribution.

Let $\boldsymbol{\phi}=(\phi_1,\dots,\phi_D)$ denote the full relevance vector, and let $z_i \in \{1,\dots,K\}$ be the latent cluster label for observation $i$. For each feature dimension $x_{ij}$, we define the conditional log-density as
\begin{equation}
\log p(x_{ij}\mid z_i=k,\phi_j)
=
\phi_j \log \mathcal{N}_{kj}
+
(1-\phi_j)\log \mathcal{N}_{0j},
\label{eq:feature_gate}
\end{equation}
where
\[
\mathcal{N}_{kj} \triangleq \mathcal{N}(x_{ij}\mid \mu_{kj},\sigma_{kj}^2),
\qquad
\mathcal{N}_{0j} \triangleq \mathcal{N}(x_{ij}\mid \mu_{0j},\sigma_{0j}^2).
\]
This formulation plays the role of a feature-gating mechanism analogous in spirit to automatic relevance determination (ARD), in the sense that feature-specific inclusion parameters control whether individual input dimensions contribute meaningfully to cluster-specific modeling \citep{neal1996bnn,tipping2001rvm}.

Assuming conditional independence across coordinates, the gated component log-density for sample $x_i$ under cluster $k$ is
\begin{equation}
\ell_{ik}(\boldsymbol{\phi})
\triangleq
\sum_{j=1}^D \log p(x_{ij}\mid z_i=k,\phi_j).
\label{eq:gated_component_logdensity}
\end{equation}
Marginalizing over the latent cluster assignment yields
\begin{equation}
\log p(x_i \mid \Theta,\boldsymbol{\phi})
=
\log \sum_{k=1}^K \pi_k \exp\!\big(\ell_{ik}(\boldsymbol{\phi})\big),
\label{eq:marginal_likelihood}
\end{equation}
where $\Theta=\{\mu,\sigma^2,\pi\}$ denotes the collection of mixture means, variances, and weights.

For optimization, we relax $\phi_j$ to a continuous variable $\tilde{\phi}_j \in (0,1)$ using a Gumbel--Sigmoid reparameterization. For binary $\phi_j$, Eq.~\eqref{eq:feature_gate} selects either the cluster-specific or background contribution. During optimization, the relaxed variable $\tilde{\phi}_j$ is used only as a differentiable surrogate on the log-density scale rather than as a separate probabilistic mixture model.

\subsection{Data-Informed Variational Objective}
\label{subsec:training}

We place a factorized variational approximation on the global feature gates:
\begin{equation}
q(\boldsymbol{\phi})
=
\prod_{j=1}^D \mathrm{Bernoulli}\!\big(\phi_j;\sigma(\eta_j)\big),
\label{eq:qphi}
\end{equation}
where $\eta_j$ are learnable variational logits. Let $\rho_j$ denote the prior inclusion probability, so that
\[
p(\phi_j)=\mathrm{Bernoulli}(\rho_j),
\qquad
q_\eta(\phi_j=1)=\sigma(\eta_j).
\]
The prior probabilities $\rho_j$ are initialized using data-informed discriminability statistics computed before training.

Since the mixture parameters $\Theta=\{\mu,\sigma^2,\pi\}$ are optimized as point estimates while $\boldsymbol{\phi}$ is treated variationally, DIVI uses a hybrid variational objective rather than a full mean-field posterior over all latent quantities. Specifically, we optimize the scaled variational objective
\begin{equation}
\mathcal{J}
=
-\sum_{i=1}^N \mathbb{E}_{q(\boldsymbol{\phi})}
\big[\log p(x_i\mid \Theta,\boldsymbol{\phi})\big]
+
\beta \sum_{j=1}^D
\mathrm{KL}\!\left(q(\phi_j)\,\|\,p(\phi_j)\right).
\label{eq:objective}
\end{equation}
We set $\beta=N$ so that the regularization term scales commensurately with the data log-likelihood $\mathcal{O}(N)$, reducing the tendency of prior influence to vanish as the sample size increases.

Crucially, we do not introduce an explicit factor $q(\mathbf Z)$. Instead, cluster assignments are handled implicitly through the marginal mixture likelihood in Eq.~\eqref{eq:marginal_likelihood}. Accordingly, the variational approximation is applied only to the global feature gates, while the mixture parameters are optimized directly and cluster assignments remain implicit in the log-sum-exp aggregation across components. Thus, Eq.~\eqref{eq:objective} should be interpreted as a scaled variational objective on $\boldsymbol{\phi}$ rather than as a joint ELBO over $(\mathbf Z,\boldsymbol{\phi})$.

In practice, we approximate the expectation in Eq.~\eqref{eq:objective} using a one-sample Monte Carlo estimator with a temperature-$T$ Gumbel--Sigmoid relaxation. Let $u_j \sim \mathrm{Uniform}(0,1)$ and $g_j=-\log(-\log u_j)$ be standard Gumbel noise. The relaxed feature gate is
\begin{equation}
\tilde{\phi}_j
=
\sigma\!\left(\frac{\eta_j+g_j}{T}\right).
\label{eq:gumbel_sigmoid}
\end{equation}
The expectation term is then approximated by substituting $\tilde{\boldsymbol{\phi}}$ into Eq.~\eqref{eq:marginal_likelihood}. This yields a differentiable objective that balances data fit against a sparsity-inducing KL regularizer on the feature gates.

\subsection{Dynamic Structure Learning via NLL Diagnostics}
\label{subsec:Dynamic}

Instead of fixing the number of clusters in advance, DIVI adopts a constructive split-based growth mechanism inspired by incremental splitting strategies. Training begins with a single component ($K=1$), and model complexity is expanded only when local fit diagnostics indicate underfit. Because the current formulation is split-only, the number of components grows monotonically during training.

Every $T_{\mathrm{split}}$ epochs, we form hard assignments using the current relaxed feature gates:
\begin{equation}
\hat z_i
=
\arg\max_k \ell_{ik}(\tilde{\boldsymbol{\phi}}).
\label{eq:hard_assignment}
\end{equation}
For each cluster $k$, let $\mathcal{I}_k=\{i:\hat z_i=k\}$ denote the assigned subset. We then define the cluster-level diagnostic score
\begin{equation}
s_k
=
-\frac{1}{|\mathcal{I}_k|}
\sum_{i\in\mathcal{I}_k}
\ell_{ik}(\tilde{\boldsymbol{\phi}}),
\label{eq:cluster_diag}
\end{equation}
that is, the average negative log-likelihood under the currently selected component. If the worst-fitting cluster satisfies
\[
\max_k s_k > \tau,
\]
a split is triggered. The target component is duplicated and its means are perturbed, while the associated log-variances and feature-gating parameters are inherited. This design allows the model to expand only when local fit remains poor under the current structure.

As a default calibration, we derive $\tau$ from the theoretical entropy of a standard multivariate Gaussian with $\sigma^2=1$ and dimension $D$. This provides a dimension-aware baseline threshold, while sensitivity around this default is examined empirically in the Results section. The full training procedure is summarized in Algorithm~\ref{alg:divi}.

\paragraph{Implementation details}
DIVI is implemented in PyTorch and optimized using Adam~\citep{kingma2014adam} with a default learning rate of $10^{-2}$, unless otherwise noted in the sensitivity analyses. Prior to training, all input features are standardized to zero mean and unit variance so that the background distribution and the default split threshold operate on a common scale. The Gumbel--Sigmoid relaxation is annealed from temperature $1.0$ to $0.1$ during training.

\begin{algorithm}[p]
\caption{DIVI: data-informed variational clustering with split-only structure growth}
\label{alg:divi}
\begin{algorithmic}[1]
\STATE \textbf{Input:} $X\in\mathbb{R}^{N\times D}$, split interval $T_{\mathrm{split}}$, max epochs $E$, threshold $\tau$.
\STATE \textbf{Output:} hard assignments $\hat Z$, feature relevance $q(\boldsymbol{\phi})$.

\STATE \textbf{Prior initialization (Mode 1):}
compute per-feature discriminability scores using a combination of KW and LLR statistics; map the normalized score to $\rho_j$ through a logistic transform.
\STATE Initialize $K \leftarrow 1$, $\mu_1 \leftarrow \bar x$, $\eta \leftarrow \mathrm{logit}(\rho)$, and $T \leftarrow T_0$.

\FOR{$t=1$ to $E$}
    \STATE Sample relaxed mask $\tilde{\boldsymbol{\phi}} \sim q_T(\boldsymbol{\phi})$ using the Gumbel--Sigmoid relaxation.
    \STATE Update $(\Theta,\eta)$ by Adam on $\mathcal{J}(\Theta,\eta)$ in Eq.~\eqref{eq:objective}; anneal $T \leftarrow \max(T_{\min},\gamma T)$.
    \IF{$t \bmod T_{\mathrm{split}}=0$}
        \STATE Compute hard assignments $\hat z_i \leftarrow \arg\max_k \ell_{ik}(\tilde{\boldsymbol{\phi}})$ and cluster subsets $\hat C_k=\{i:\hat z_i=k\}$.
        \STATE Compute diagnostic scores $S_k \leftarrow -|\hat C_k|^{-1}\sum_{i\in\hat C_k}\ell_{ik}(\tilde{\boldsymbol{\phi}})$ and select $k^\star \leftarrow \arg\max_k S_k$.
        \IF{$S_{k^\star}>\tau$}
            \STATE Split $k^\star$ by duplicating and perturbing its mean parameters; inherit the corresponding log-variances and feature-gating logits.
            \STATE Set $K \leftarrow K+1$ and reinitialize Adam.
        \ENDIF
    \ENDIF
\ENDFOR
\STATE \textbf{Return} hard assignments $\hat Z=\{\hat z_i\}$ and feature relevance $q(\boldsymbol{\phi})=\sigma(\eta)$.
\end{algorithmic}
\end{algorithm}

\paragraph{Computational complexity}
Under the current implementation, the dominant cost of DIVI is the evaluation of gated Gaussian log-densities for all samples, components, and features. For data $X \in \mathbb{R}^{N\times D}$ and $K$ active components, each training epoch requires $O(NKD)$ operations, while the KL term on the global feature gates contributes only $O(D)$. The one-time data-informed initialization costs approximately $O(I_A N D K_0 + ND)$, where $K_0$ is the rough clustering size and $I_A$ is the number of rough $k$-means iterations. Split diagnostics are performed every $T_{\mathrm{split}}$ epochs and require one additional forward pass, so the total training cost is
\[
O\!\left(\sum_{t=1}^{E} N D K_t \right),
\]
which is upper bounded by $O(E N D K_{\max})$ under split-only growth. Parameter storage is $O(KD + D)$, whereas the dominant working memory during backpropagation scales on the order of $O(NKD)$ under the current implementation.

\section{Experimental Design}
\label{subsec:setup}

Our empirical study is organized to examine five aspects of DIVI: mechanism verification under severe feature noise, robustness to model misspecification, interpretability of the learned feature relevance, practical computational behavior as the data scale increases, and sensitivity to the main structural and regularization parameters.

The design includes one matched synthetic benchmark, two misspecified synthetic variants, and three real-world datasets. Unless otherwise noted, all inputs are standardized before fitting. This is particularly important for DIVI because the split criterion and the background distribution are both calibrated relative to standardized feature scales.

The matched synthetic benchmark considers a three-cluster setting with $(N \in \{200,1000\}, D=100)$, where the first $10$ dimensions are informative and the remaining $90$ are nuisance. For the informative coordinates, cluster-specific means are set to $-2$, $0$, and $2$, respectively, while the nuisance coordinates are generated as independent Gaussian noise with standard deviation $\sigma=3.0$. Because both the true partition and the informative feature subset are known, this benchmark serves as the primary setting for mechanism verification and for evaluating informative-support recovery.

To examine robustness beyond the matched Gaussian setting, we further construct two misspecified synthetic variants with the same overall difficulty level $(K=3,\;D=100,\;10\% \text{ informative},\;90\% \text{ nuisance})$. In the \emph{heavy-tailed signal} setting, the informative dimensions are generated from a Student-\(t\) distribution with 5 degrees of freedom, standardized to unit variance before applying the cluster-specific mean shifts, while the nuisance dimensions remain independent Gaussian noise with $\sigma=3.0$. In the \emph{correlated-noise} setting, the informative coordinates remain Gaussian, but the nuisance dimensions are generated from block-correlated Gaussian noise with within-block correlation $\rho=0.6$, block size $10$, and marginal scale $\sigma=3.0$. These variants are intended to test whether the main empirical behavior of DIVI persists when the data-generating mechanism departs from the model assumptions.

For real-data experiments, we consider the UCI Wine dataset \citep{aeberhard1994wine} $(N=178, D=13)$ as a small-scale interpretability example, since the learned relevance profile can be compared against known chemically discriminative variables. We also use an ISOLET subset \citep{uci_isolet} $(N=1560, D=617)$ consisting of the first five spoken letters (A--E), which provides a structured high-dimensional spectral benchmark. Finally, we evaluate DIVI on a subset of 20 Newsgroups (20NG) \citep{lang1995newsweeder} $(N=2000, D=384)$ constructed from four categories: \texttt{sci.space}, \texttt{rec.autos}, \texttt{talk.politics.mideast}, and \texttt{comp.graphics}. Raw documents are encoded into dense sentence embeddings using the pre-trained \texttt{all-MiniLM} \texttt{-L6-v2} Sentence-BERT model~\citep{Reimers2019SentenceBERTSE}. The resulting embeddings are \(\ell_2\)-normalized and then standardized before clustering. This benchmark is included to assess DIVI on modern dense semantic representations, where relevant information is expected to be distributed across many embedding dimensions.

Our external baselines are K-means with oracle $K$, diagonal-covariance GMM with oracle $K$, and Sparse K-means (SPKM)~\citep{witten2010framework}. Providing oracle $K$ for K-means and GMM isolates the effect of feature noise from the separate issue of model-order selection. SPKM serves as the main feature-selective clustering comparator based on $\ell_1$-penalized feature weights. For the misspecified synthetic robustness experiments, we report oracle-$K$ K-means and oracle-$K$ GMM as the primary external comparators, since the goal there is to isolate the effect of distributional misspecification rather than to revisit every feature-selective baseline.

In addition, we consider two internal ablations: \textbf{DIVI-NonInfo} (Mode 2), which removes data-informed guidance by setting $\rho_j=0.5$, and \textbf{DIVI-Random} (Mode 3), which initializes $\rho_j \sim \mathrm{Uniform}(0,1)$. These ablations are used primarily in the synthetic experiments, where the role of prior initialization can be isolated most directly. Full ablation summaries are reported in Appendix B. The main real-data results focus on the proposed \textbf{DIVI-Info} (Mode 1).

For evaluation, we report \textbf{ARI} and \textbf{NMI} as the primary clustering metrics throughout. On synthetic data, where the true informative subset is known, we additionally report the \textbf{Feature F1-score} to quantify recovery of the informative support. For computational analyses, we further summarize wall-clock time, the final discovered number of clusters, and the number of active dimensions retained by DIVI.

Because DIVI uses split-only structure growth, its behavior under misspecification depends on the split schedule. For the misspecified robustness experiments, we therefore fix the sensitivity-selected setting $(\tau_{\mathrm{mult}}, T_{\mathrm{split}})=(1.00,120)$ and apply it uniformly across all misspecified synthetic settings. This schedule was chosen in a separate sensitivity analysis to reduce the mild over-splitting observed under the default configuration while leaving the remaining optimization settings unchanged. Complete hyperparameter settings for all experiments are reported in Appendix B.

\section{Results}

\subsection{Matched Synthetic Benchmark}
\label{sec:synthetic_main}

We first consider the matched synthetic benchmark, which provides a controlled setting for examining how DIVI behaves under severe feature contamination when both the true partition and the informative support are known. This allows us to evaluate not only partition quality but also recovery of the informative subspace. For this benchmark, we use the current DIVI implementation together with the sensitivity-selected split interval \(T_{\mathrm{split}}=120\), which avoids the mild over-splitting observed under the default split schedule in this split-only formulation.

\begin{table}[t]
\centering
\footnotesize
\setlength{\tabcolsep}{3.8pt}
\renewcommand{\arraystretch}{0.97}
\caption{Clustering performance under high-dimensional noise. Results are averaged over 20 independently generated synthetic datasets with $D=100$ and 90\% irrelevant dimensions. DIVI is evaluated using the current implementation with the sensitivity-selected split interval $T_{\mathrm{split}}=120$. Feature F1 is reported only for methods with explicit feature selection or feature gating.}
\label{tab:synthetic_main}
\begin{tabular}{lcccccccc}
\toprule
& \multicolumn{4}{c}{$N=200$ (Small-$n$)} & \multicolumn{4}{c}{$N=1000$ (Large-$n$)} \\
\cmidrule(lr){2-5} \cmidrule(lr){6-9}
Method & $K$ & ARI & NMI & F1 & $K$ & ARI & NMI & F1 \\
\midrule
K-means
& 3.000 & .909 & .885 & \multicolumn{1}{c}{--}
& 3.000 & .992 & .985 & \multicolumn{1}{c}{--} \\
& (.000) & (.048) & (.053) & \multicolumn{1}{c}{--}
& (.000) & (.006) & (.010) & \multicolumn{1}{c}{--} \\
\addlinespace[1pt]

GMM
& 3.000 & .992 & .987 & \multicolumn{1}{c}{--}
& 3.000 & .995 & .991 & \multicolumn{1}{c}{--} \\
& (.000) & (.010) & (.016) & \multicolumn{1}{c}{--}
& (.000) & (.004) & (.007) & \multicolumn{1}{c}{--} \\
\addlinespace[1pt]

SPKM
& 3.000 & .993 & .989 & .184
& 3.000 & .996 & .993 & .190 \\
& (.000) & (.010) & (.016) & (.002)
& (.000) & (.003) & (.006) & (.003) \\
\addlinespace[1pt]

DIVI-Info
& 3.000 & .990 & .985 & .772
& 3.000 & .995 & .990 & .990 \\
& (.000) & (.011) & (.017) & (.099)
& (.000) & (.004) & (.007) & (.019) \\
\addlinespace[1pt]

DIVI-NonInfo
& 3.000 & .868 & .878 & .182
& 3.000 & .947 & .951 & .182 \\
& (.000) & (.205) & (.166) & (.000)
& (.000) & (.131) & (.102) & (.000) \\
\addlinespace[1pt]

DIVI-Random
& 3.000 & .700 & .728 & .186
& 3.000 & .896 & .893 & .186 \\
& (.000) & (.235) & (.191) & (.027)
& (.000) & (.155) & (.130) & (.027) \\
\bottomrule
\end{tabular}
\end{table}

Table~\ref{tab:synthetic_main} summarizes the main synthetic results under two sample-size regimes with 90\% irrelevant dimensions. Under the current protocol, the oracle-\(K\) external baselines are substantially stronger than in our earlier conference-style experiments, so this benchmark should not be interpreted as a setting in which generic baselines fail categorically. Instead, its main value is to isolate the role of feature relevance learning and prior-guided initialization under controlled high-dimensional noise.

In this setting, DIVI-Info remains highly competitive in clustering quality while showing a clear advantage over the uninformed ablations in informative-support recovery. In particular, informative prior initialization leads to substantially stronger feature-level F1 than either DIVI-NonInfo or DIVI-Random at both sample sizes, indicating that prior guidance is an important component for stabilizing feature gating and recovering the informative subspace under overwhelming noise.

The comparison with the external baselines should therefore be interpreted carefully. K-means, diagonal-covariance GMM, and SPKM all achieve strong partition recovery on this matched design. However, these methods do not provide the same kind of probabilistic feature-gating mechanism, and SPKM in particular achieves high clustering accuracy while retaining substantially weaker support recovery than DIVI-Info. This contrast is especially informative because it shows that strong partition recovery does not by itself imply accurate recovery of the informative subspace. The principal value of DIVI-Info on this benchmark therefore lies not in universal dominance in flat clustering accuracy alone, but in combining competitive partition recovery with interpretable feature gating and substantially stronger informative-support recovery.

\subsection{Robustness to Model Misspecification}

We next examine whether the qualitative behavior observed on the matched synthetic benchmark persists when the data-generating mechanism departs from the feature-gated Gaussian structure assumed by DIVI. To this end, we consider two misspecified variants that preserve the same overall difficulty level ($D=100$, 10 informative dimensions, and 90\% nuisance dimensions): \emph{heavy-tailed signal}, where the informative coordinates are non-Gaussian, and \emph{correlated noise}, where nuisance dimensions exhibit block dependence. Because DIVI uses split-only structure growth, we fix the robustness experiments to the sensitivity-selected split schedule $(\tau_{\mathrm{mult}}, T_{\mathrm{split}})=(1.00,120)$, which removes the mild over-splitting observed under the default schedule and consistently recovers the true $K=3$ across these settings. We do not repeat SPKM in the main misspecification table, because the purpose here is to isolate robustness to distributional misspecification relative to standard oracle-\(K\) baselines rather than to revisit the full set of feature-selective comparators.

Table~\ref{tab:misspec_robustness_main} shows the main comparison against oracle-\(K\) K-means and oracle-\(K\) GMM. Under heavy-tailed signal, DIVI-Info maintained high clustering quality while support recovery improved markedly with sample size. In particular, Feature F1 increased from $0.768$ at $N=200$ to $0.990$ at $N=1000$, whereas clustering accuracy remained high but did not improve monotonically under Gaussian model misspecification. This pattern suggests that, under heavy-tailed signal contamination, larger sample size primarily strengthens informative-support recovery, while the clustering metric itself is already near saturation.

Under correlated noise, exact support recovery became substantially harder for all methods, but DIVI-Info still showed the strongest clustering robustness among the methods reported in the main table, outperforming both oracle K-means and oracle GMM in ARI at both sample sizes. The remaining gap between ARI and Feature F1 is also informative: even when exact support recovery is difficult under block-dependent nuisance structure, the feature-gating mechanism can still attenuate irrelevant dimensions sufficiently to improve partition recovery. Full ablation results for DIVI-NonInfo and DIVI-Random are deferred to Appendix B, where they confirm the same qualitative pattern: without data-informed initialization, clustering may remain partially recoverable, but reliable identification of the informative subspace deteriorates markedly.

\begin{table}[t]
\centering
\footnotesize
\setlength{\tabcolsep}{4pt}
\caption{Main robustness results under model misspecification. DIVI-Info was evaluated with the sensitivity-selected split schedule $(\tau_{\mathrm{mult}}, T_{\mathrm{split}})=(1.00,120)$. Values are mean (std) over 20 independently generated datasets.}
\label{tab:misspec_robustness_main}
\setlength{\tabcolsep}{4pt}
\begin{tabular}{llcccc}
\toprule
Scenario & $N$ & K-means ARI & GMM ARI & DIVI-Info ARI & DIVI-Info F1 \\
\midrule
Heavy-tailed signal & 200  & 0.913 (0.034) & 0.853 (0.238) & \textbf{0.989} (0.015) & \textbf{0.768} (0.095) \\
Heavy-tailed signal & 1000 & \textbf{0.986} (0.005) & 0.971 (0.097) & 0.964 (0.107) & \textbf{0.990} (0.020) \\
Correlated noise    & 200  & 0.283 (0.080) & 0.307 (0.176) & \textbf{0.419} (0.153) & \textbf{0.243} (0.038) \\
Correlated noise    & 1000 & 0.381 (0.037) & 0.497 (0.007) & \textbf{0.712} (0.261) & \textbf{0.254} (0.031) \\
\bottomrule
\end{tabular}
\end{table}

\subsection{Real-Data Behavior and Interpretability}

We then turn to real-data experiments to examine how the feature-gating mechanism behaves when the signal is less sparse, more broadly distributed across dimensions, and no longer aligned with a simple synthetic support structure. On dense representations such as 20NG embeddings, DIVI does not behave as an aggressively sparse selector. Instead, it acts more like an adaptive filter that preserves most information-bearing dimensions while attenuating a relatively small non-informative tail. In this setting, K-means remains both faster and stronger in raw flat clustering accuracy, but DIVI still provides a meaningful weighted representation together with an adaptive structural summary. We therefore interpret the 20NG results not as evidence of runtime or accuracy superiority, but as a useful case in which feature gating remains stable and yields an interpretable weighted subspace on dense semantic embeddings.

A complementary perspective is provided by the Wine data, which we treat as a small-scale interpretability illustration rather than as a primary benchmark. In this example, the learned relevance profile aligns reasonably well with known chemical differences among wine categories and highlights a compact subset of variables that contributes most strongly to cluster separation. The main value of this experiment is therefore qualitative: it shows that the feature-gating mechanism can recover a meaningful low-dimensional view in a fully unsupervised setting, even though the overall framework is optimized for clustering rather than for supervised variable selection.

\begin{figure}[!t]
  \centering
  \includegraphics[width=0.7\columnwidth]{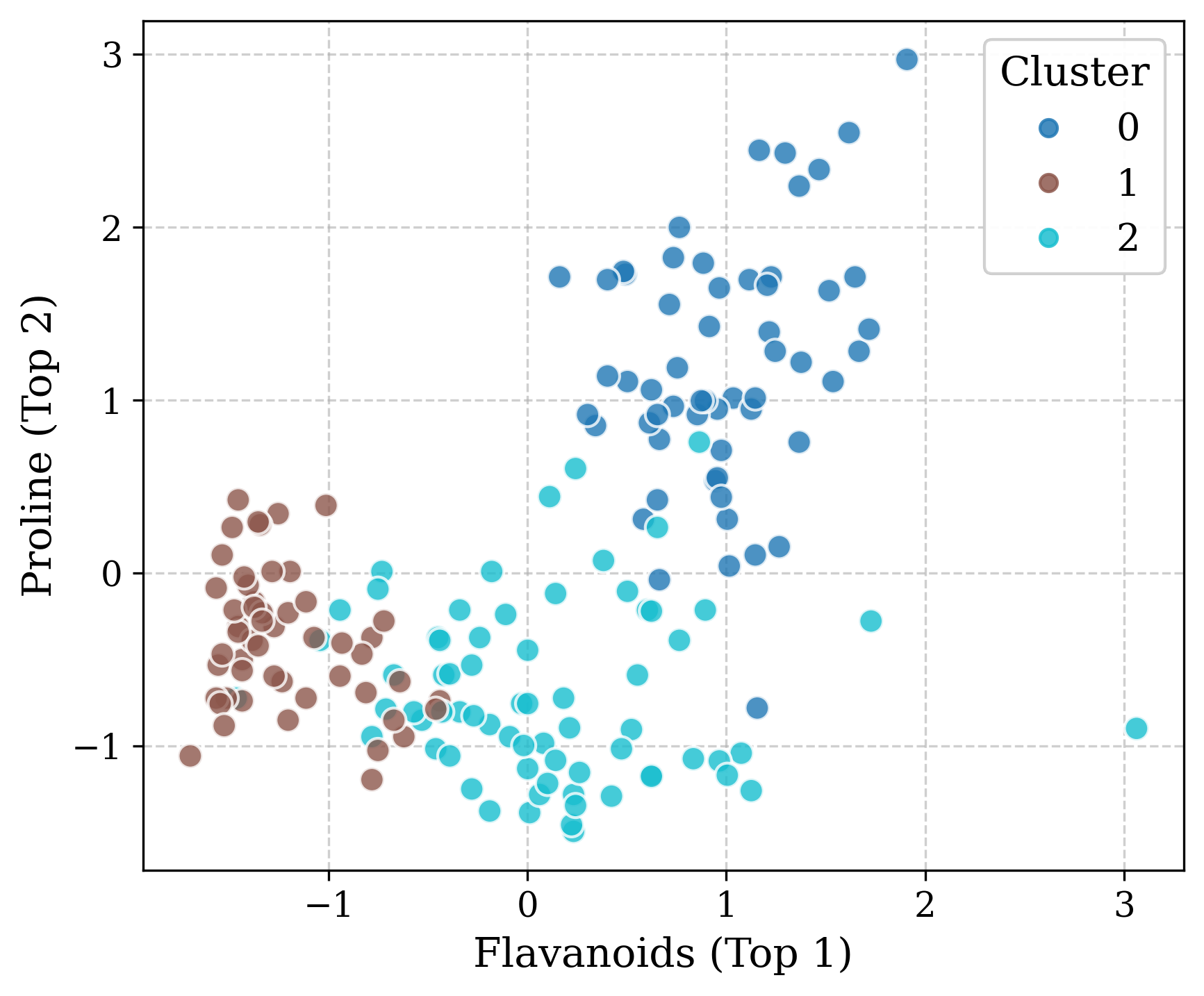}
  \caption{Interpretability illustration on the UCI Wine dataset. The observations are visualized in the two most relevant dimensions identified by the learned feature-gating mechanism. Although this experiment is small in scale, it shows that the learned relevance profile can produce a meaningful low-dimensional view that aligns reasonably well with known chemical differences among wine categories.}
  \label{fig:wine_scatter}
\end{figure}

Figure~\ref{fig:wine_scatter} provides a simple interpretability illustration on the Wine dataset by visualizing the observations in the two most relevant dimensions identified by the learned feature-gating mechanism.

ISOLET provides a more conservative and less favorable real-data case. Unlike the dense semantic representation of 20NG, the discriminative information in ISOLET is distributed across many spectral dimensions, so broad feature retention is not itself unexpected. However, under the current split-and-gating scheme, DIVI remains conservative on this dataset: it retains a large active subspace while stopping at a slightly under-expanded cluster structure. We therefore treat ISOLET as a useful limitation case rather than as a clear success story. More broadly, it suggests that the present formulation can behave conservatively on structured high-dimensional signals when cluster expansion and feature gating must be learned simultaneously.

\subsection{Computational Behavior and Scalability}

We next examine the computational behavior of DIVI from both practical and scaling perspectives, with the goal of clarifying how the cost of joint feature gating and adaptive structure growth changes across representative settings. The former reflect empirical cost in representative high-dimensional applications, whereas the latter isolate how runtime changes as the feature dimension \(D\) or sample size \(N\) increases. These computational summaries are consistent with the complexity characterization given in the Method section and help clarify how the practical cost of DIVI changes across representative real and synthetic settings.

\begin{table}[t]
\centering
\footnotesize
\setlength{\tabcolsep}{4pt}
\caption{Real-data clustering and computational summaries on ISOLET and 20NG embeddings, where 20NG denotes the 20 Newsgroups subset. Results are reported as mean (standard deviation) over repeated runs. For K-means, GMM, and SPKM, the true number of clusters is supplied. DIVI starts from a single component and expands via split diagnostics. Active dimensions are reported for feature-selective methods only (DIVI and SPKM).}
\label{tab:runtime-real-main}
\begin{tabular}{llccccc}
\toprule
Dataset & Method & ARI & NMI & Time (s) & Final $K$ & Active dims \\
\midrule
ISOLET & DIVI-Info & 0.460 (0.143) & 0.585 (0.122) & 22.26 (1.14) & 4.0 (0.0) & 578.0 (1.7) \\
ISOLET & K-means    & 0.632 (0.001) & 0.716 (0.001) & 0.59 (0.04)  & 5.0 (0.0) & -- \\
ISOLET & GMM       & 0.580 (0.112) & 0.684 (0.068) & 2.81 (0.75)  & 5.0 (0.0) & -- \\
ISOLET & SPKM      & 0.551 (0.000) & 0.640 (0.000) & 9.26 (0.66)  & 5.0 (0.0) & 23.0 (0.0) \\
\midrule
20NG   & DIVI-Info & 0.759 (0.133) & 0.753 (0.074) & 16.49 (1.03) & 4.0 (0.0) & 351.6 (1.1) \\
20NG   & K-means    & 0.845 (0.002) & 0.803 (0.002) & 0.40 (0.02)  & 4.0 (0.0) & -- \\
20NG   & GMM       & 0.718 (0.167) & 0.735 (0.088) & 1.42 (0.15)  & 4.0 (0.0) & -- \\
20NG   & SPKM      & 0.514 (0.000) & 0.515 (0.000) & 7.98 (0.61)  & 4.0 (0.0) & 30.0 (0.0) \\
\bottomrule
\end{tabular}
\end{table}

Table~\ref{tab:runtime-real-main} reports real-data clustering and runtime summaries on ISOLET and 20NG embeddings. These results are not intended to demonstrate raw speed superiority: fixed-$K$ baselines remain substantially faster. Rather, the table shows that DIVI remains computationally feasible on moderately large high-dimensional datasets, albeit with a clear runtime premium due to joint feature gating and split-based structure adaptation.

\begin{figure}[!t]
  \centering
  \includegraphics[width=0.75\columnwidth]{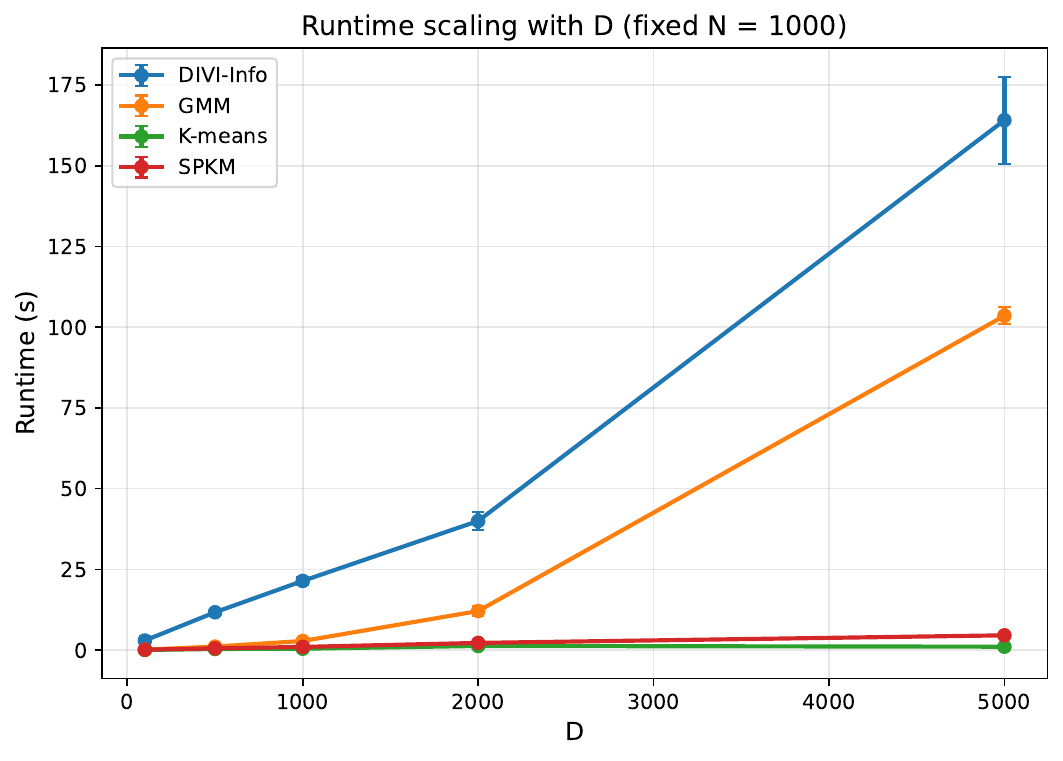}
  \caption{Runtime scaling with feature dimension $D$ on synthetic data under fixed sample size ($N=1000$). DIVI-Info exhibits a monotone increase in wall-clock cost as dimensionality grows, reflecting the increasing cost of feature-wise variational optimization.}
  \label{fig:runtime_dscale}
\end{figure}

Figure~\ref{fig:runtime_dscale} shows runtime as a function of \(D\) under fixed sample size. As dimensionality increases, the runtime of DIVI rises monotonically, reflecting the increasing cost of feature-wise variational optimization. A practically important exception appears at the most extreme regime, where clustering quality deteriorates together with a sharp increase in active dimensions. This indicates a high-dimensional failure regime under fixed regularization, rather than a simple benign slowdown.

\begin{figure}[!t]
  \centering
  \includegraphics[width=0.75\columnwidth]{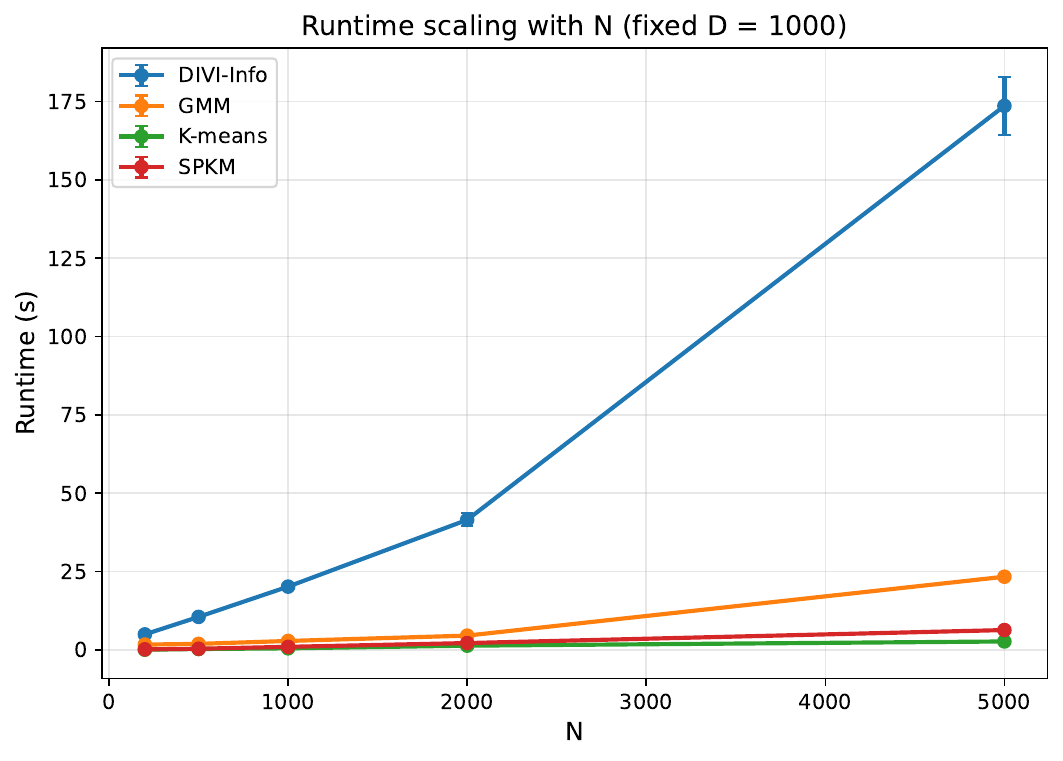}
  \caption{Runtime scaling with sample size $N$ on synthetic data under fixed dimensionality ($D=1000$). Runtime increases with sample size for all methods, with DIVI-Info incurring the highest computational cost due to joint feature gating and split-based structure adaptation.}
  \label{fig:runtime_nscale}
\end{figure}

Figure~\ref{fig:runtime_nscale} shows runtime as a function of \(N\) under fixed dimensionality. Runtime again increases predictably with sample size, but in contrast to the high-\(D\) failure regime, clustering performance improves as more observations become available. This pattern suggests that larger sample sizes stabilize the feature-gating mechanism and support more reliable adaptive clustering, albeit at an increased computational cost.

Overall, the runtime results position DIVI as a computationally heavier but still practical alternative to fixed-\(K\) baselines. Its additional cost should be understood as the price of jointly performing feature gating and split-based structure adaptation, rather than as unnecessary optimization overhead.

These computational trends already suggest that the main tuning parameters of DIVI should be interpreted structurally rather than as generic black-box hyperparameters, since they directly govern the aggressiveness of split-based growth and the strength of feature gating. In particular, the cost of adaptive structure growth and the stability of feature gating motivate the targeted sensitivity analyses reported next.

\subsection{Sensitivity Analysis}

We then investigate parameter sensitivity in order to distinguish which aspects of DIVI reflect stable structural behavior and which depend more directly on the tuning of split frequency, feature regularization, and optimization settings. In particular, the split interval \(T_{\text{split}}\) governs how frequently local underfit is checked, the KL scaling factor \(\beta\) controls the strength of feature regularization, and the threshold parameter \(\tau\) determines when a poorly fitted cluster should trigger expansion. We therefore examine sensitivity not merely as a robustness checklist, but as a way to understand how these parameters shape adaptive structure growth and feature gating in high-dimensional noise.

Among all tuning parameters, \(T_{\text{split}}\) exhibits the strongest structural effect. When split checks are performed too frequently, DIVI tends to over-expand the model, producing inflated final cluster counts, longer runtime, and substantially worse ARI/NMI. This behavior is consistent with the split-only design of the method: once an unnecessary split is introduced, the model cannot merge back to a simpler structure. In contrast, more conservative split intervals yield a markedly better tradeoff between clustering accuracy and computational cost. These results indicate that \(T_{\text{split}}\) is the most critical practical parameter for controlling irreversible structural growth.

\begin{figure}[htbp]
  \centering
  \includegraphics[width=0.75\columnwidth]{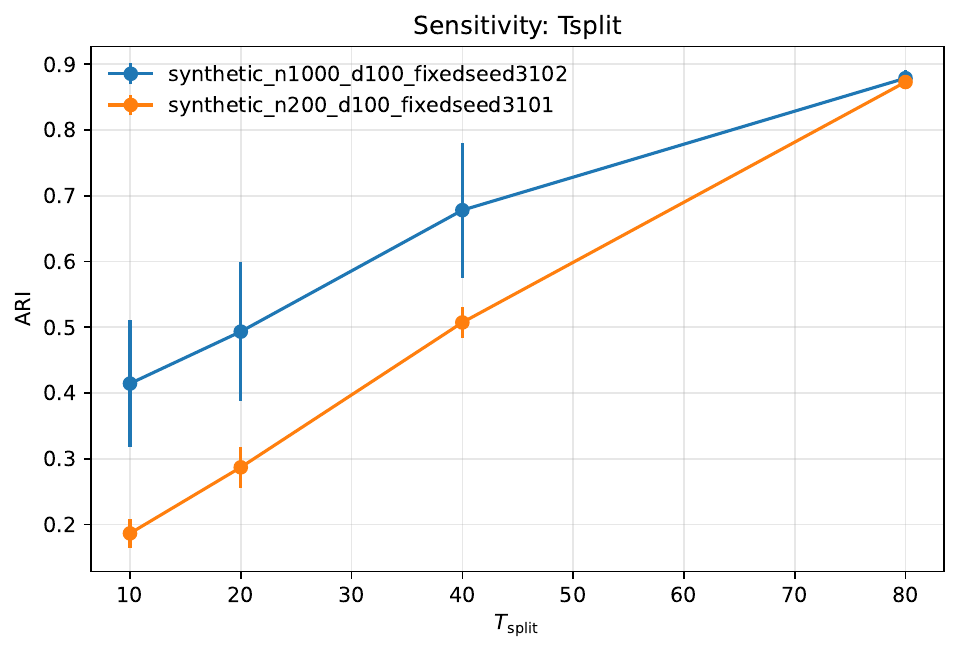}
  \caption{Sensitivity of DIVI to the split interval \(T_{\mathrm{split}}\) on fixed synthetic datasets, measured by ARI. Very small split intervals lead to substantially worse clustering performance, whereas more conservative split intervals yield markedly better ARI across both sample-size regimes.}
  \label{fig:sens_tsplit}
\end{figure}

The role of \(\beta\) is qualitatively different. Across a broad range of values, ARI and NMI remain comparatively stable, whereas feature-level recovery and dimensional parsimony change much more substantially. As \(\beta\) increases, DIVI retains fewer dimensions and produces cleaner recovery of the informative subspace, while the clustering partition itself is only mildly affected. This suggests that \(\beta\) primarily regulates feature gating rather than cluster assignment accuracy, and it provides practical support for the default scaling choice \(\beta = N\), which places the data-fit and regularization terms on a comparable scale.

\begin{figure}[htbp]
  \centering
  \includegraphics[width=0.75\columnwidth]{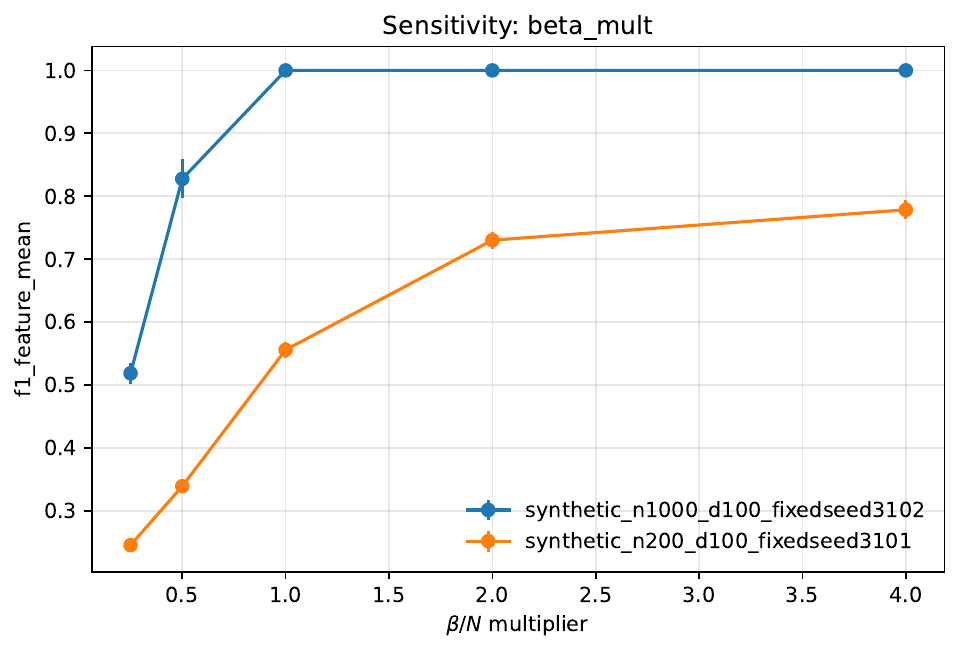}
  \caption{Sensitivity of DIVI to the KL-scaling multiplier \(\beta/N\) on fixed synthetic datasets, measured by Feature F1. Larger \(\beta/N\) improves or stabilizes informative-support recovery, with the gains being most pronounced in the small-\(n\) setting. The results support \(\beta = N\) as a practical default for feature-gating regularization.}
  \label{fig:sens_beta}
\end{figure}

By contrast, moderate perturbations around the default entropy-based threshold \(\tau\) produce nearly unchanged clustering, feature, and runtime summaries. This indicates that the split criterion is locally robust in a neighborhood of the default value and is therefore less sensitive than either \(T_{\text{split}}\) or \(\beta\) in practice. For this reason, we treat \(\tau\) as a secondary tuning parameter and defer its full sensitivity table to Table~\ref{tab:sens_tau} in Appendix B.

We further examined two optimization-related quantities: the learning rate and the temperature annealing endpoint of the Gumbel--Sigmoid relaxation. These analyses serve a different purpose from the structural sensitivity results above. Rather than changing the statistical role of the model, they assess whether the main qualitative patterns are driven primarily by optimizer-specific choices. In our experiments, moderate changes in learning rate and annealing endpoint do not materially change the main qualitative conclusions: the feature-gating mechanism remains stable over a reasonable neighborhood of the default settings, and the dominant sensitivities continue to be associated with structural growth and feature regularization rather than with low-level optimizer choices.

Taken together, the sensitivity results reveal a useful division of labor among the main tuning parameters. \(T_{\text{split}}\) controls the aggressiveness of adaptive expansion, \(\beta\) controls feature parsimony, \(\tau\) is locally stable around its default value, and the optimization-related settings mainly affect robustness rather than the qualitative form of the solution. This separation is practically useful because it allows tuning decisions to be interpreted in structural terms rather than by trial and error alone.

Detailed sensitivity summaries for \(T_{\mathrm{split}}\), \(\beta\), \(\tau\), the learning rate, and the temperature annealing endpoint are provided in Tables~\ref{tab:sens_tsplit}--\ref{tab:sens_temp_end} in Appendix B.

\subsection{Failure Regimes and Practical Scope}

Finally, we synthesize the preceding empirical results to clarify the regimes in which DIVI behaves reliably, the settings in which it becomes conservative or unstable, and the practical limits of the current split-and-gating formulation. Because the current method uses split-only structure growth, its structural behavior is most sensitive to the split schedule. When split checks are performed too frequently, the model may over-expand irreversibly; when they are too infrequent, it may remain under-expanded. This is not a generic optimization failure, but a direct consequence of one-way structural adaptation without merge operations. Our sensitivity analysis therefore suggests interpreting $T_{\mathrm{split}}$ as the primary control for structural aggressiveness rather than as an incidental hyperparameter: too-frequent split checks promote irreversible over-expansion, whereas overly conservative schedules can leave the model under-expanded.

Second, model misspecification affects clustering recovery and support recovery differently. Under heavy-tailed signal, informative-support recovery improves substantially with sample size, while clustering accuracy remains high but does not improve monotonically under Gaussian misspecification. Under correlated nuisance dependence, DIVI retains a clear clustering advantage over the compared oracle baselines, but exact recovery of the informative support becomes substantially harder. These results indicate that robustness in partition recovery transfers more readily than exact support recovery once the nuisance structure departs from the independent-noise setting.

Third, on dense real-world representations such as ISOLET and 20NG embeddings, DIVI should not be interpreted as an aggressively sparse selector. Instead, it behaves more like an adaptive filter that preserves broadly distributed signal while attenuating a smaller non-informative tail. In such regimes, the current split-and-gating formulation may remain conservative in structure growth and can enter a high-dimensional failure regime in which runtime inflation is accompanied by deterioration in clustering quality. We therefore view the present formulation as most reliable in settings where adaptive feature weighting and moderate structure growth are both beneficial, but less ideal when the signal is highly distributed or when reversible structural adaptation would be required.

\section{Discussion and Limitations}

The current empirical results suggest that DIVI is best understood not as a universally dominant clustering algorithm, but as a structured variational framework for noisy high-dimensional clustering with interpretable feature gating and adaptive growth. Its main contribution lies in coupling three components that are often treated separately: feature relevance learning, cluster-structure adaptation, and data-informed initialization for optimization stabilization. From this perspective, the value of DIVI is less about raw benchmark dominance and more about providing a practically useful framework whose behavior can be examined in both statistical and computational terms.

Several strengths are consistently supported by the current experiments. On synthetic stress tests with severe feature contamination, DIVI remains competitive while explicitly recovering informative subspaces. The additional runtime and sensitivity analyses further clarify that its tuning parameters play distinct and interpretable roles rather than acting as purely opaque hyperparameters. In particular, the split interval primarily governs structural expansion, whereas the KL scaling factor mainly controls feature parsimony. These analyses strengthen the practical interpretation of DIVI and help distinguish genuine model behavior from optimizer-specific artifacts.

At the same time, the current formulation has important limitations. First, the split-only growth mechanism is irreversible, so early over-expansion cannot be corrected by later merge operations; this is consistent with the strong sensitivity observed for very small split intervals. Second, feature relevance is currently modeled globally rather than cluster-specifically, which may be restrictive in settings where different cluster pairs depend on different local subsets of variables. Third, the present formulation adopts a hybrid variational treatment in which mixture parameters are point-estimated while feature gates are handled through a relaxed posterior approximation. This improves tractability, but it also limits the probabilistic scope of the current method.

The real-data results further highlight this tradeoff between practicality and generality. On dense representations such as 20NG embeddings, DIVI behaves more like an adaptive noise filter than an aggressively sparse selector. On more challenging datasets, such as ISOLET or gene-expression benchmarks, the current split-and-gating scheme can remain conservative, yielding broad active subspaces and, in some cases, under-expanded cluster structure. We therefore view these results not as contradictions of the method, but as useful evidence about the regimes in which the present formulation is effective and the regimes in which it remains limited.

These limitations suggest several natural directions for future work. More evidence-based split criteria, richer variational formulations with explicit latent assignment distributions, and hierarchical or cluster-specific feature relevance mechanisms may provide a stronger probabilistic foundation and greater expressive flexibility. However, such extensions would substantially expand the scope of the present paper. Our goal here is more modest: to clarify the empirical behavior, computational profile, and practical boundaries of the current DIVI formulation, and to provide a more complete basis for evaluating its usefulness in noisy high-dimensional clustering problems.

\section{Conclusion}

We studied DIVI, a data-informed variational clustering framework for noisy high-dimensional data with explicit feature gating and split-based adaptive structure growth. Rather than treating feature relevance, clustering, and model complexity as disconnected tasks, DIVI couples them within a single practical optimization framework. Our experiments clarify the empirical and computational behavior of the method: DIVI remains competitive under severe feature noise, exhibits interpretable feature-gating behavior, and is computationally feasible, while also incurring nontrivial overhead relative to fixed-\(K\) baselines.

The additional runtime and sensitivity analyses further refine the practical scope of the method. In particular, they show that split frequency strongly affects irreversible structural expansion, whereas KL scaling mainly controls feature parsimony. They also reveal that the current formulation can remain conservative on some real datasets and may enter failure regimes under extreme dimensionality. Taken together, these findings suggest that DIVI is best interpreted as a practical variational framework for noisy high-dimensional clustering rather than as a universally dominant or fully Bayesian generative solution.

Several directions remain open for future work, including more evidence-based split criteria, richer variational formulations with explicit latent assignment distributions, and hierarchical or cluster-specific feature relevance mechanisms. More broadly, the present study is intended to provide a clearer empirical and computational characterization of the current DIVI formulation and a more transparent basis for evaluating its usefulness in noisy high-dimensional clustering problems.

\bibliographystyle{elsarticle-harv}
\bibliography{reference}

\clearpage
\appendix

\renewcommand{\theequation}{\Alph{section}.\arabic{equation}}
\renewcommand{\thetable}{\Alph{section}.\arabic{table}}
\renewcommand{\thefigure}{\Alph{section}.\arabic{figure}}

\section{Mathematical Details}
\label{app:math_details}

\setcounter{equation}{0}
\setcounter{table}{0}
\setcounter{figure}{0}

\subsection{Feature-Gated Mixture Construction}
\label{app:model}

DIVI uses a feature-gated Gaussian mixture in which each dimension is explained either by a cluster-specific Gaussian component or by a broad global background distribution. Let $x_i \in \mathbb{R}^D$ denote observation $i$, and let $z_i \in \{1,\dots,K\}$ denote its latent cluster label.

\paragraph{Mixture weights}
The mixture weights are parameterized by trainable logits and normalized through a softmax transform:
\[
\pi_k
=
\frac{\exp(\alpha_k)}{\sum_{\ell=1}^K \exp(\alpha_\ell)},
\qquad
k=1,\dots,K.
\]
In the implementation, the logits $\alpha_k$ are optimized directly as unconstrained parameters.

\paragraph{Feature relevance}
Each feature $j$ has a binary global relevance indicator $\phi_j \in \{0,1\}$ with prior
\[
p(\phi_j)=\mathrm{Bernoulli}(\rho_j),
\qquad
p(\boldsymbol{\phi})=\prod_{j=1}^D \mathrm{Bernoulli}(\rho_j).
\]
The prior inclusion probabilities $\rho_j$ are constructed by Step~A in Mode~1, fixed at $0.5$ in Mode~2, and sampled from $\mathrm{Uniform}(0,1)$ in Mode~3.

\paragraph{Cluster-specific and background emissions}
For cluster $k$ and feature $j$, DIVI defines the conditional log-density
\begin{equation}
\log p(x_{ij}\mid z_i=k,\phi_j)
=
\phi_j \log \mathcal{N}_{kj}
+
(1-\phi_j)\log \mathcal{N}_{0j},
\label{eq:app_gate_density}
\end{equation}
where
\[
\mathcal{N}_{kj}
=
\mathcal{N}(x_{ij}\mid \mu_{kj},\sigma_{kj}^2),
\qquad
\mathcal{N}_{0j}
=
\mathcal{N}(x_{ij}\mid \mu_{0j},\sigma_{0j}^2).
\]
The background parameters $(\mu_0,\sigma_0^2)$ are fixed and intentionally broad in the implementation. In particular, the default background mean is zero and the default background log-variance is set to a large constant.

Assuming conditional independence across dimensions, the gated component log-density for observation $x_i$ under cluster $k$ is
\begin{equation}
\ell_{ik}(\boldsymbol{\phi})
=
\sum_{j=1}^D
\log p(x_{ij}\mid z_i=k,\phi_j).
\label{eq:app_component_ll}
\end{equation}
Marginalizing over the latent cluster assignment then yields
\begin{equation}
\log p(x_i\mid \Theta,\boldsymbol{\phi})
=
\log\sum_{k=1}^K \pi_k \exp\!\big(\ell_{ik}(\boldsymbol{\phi})\big),
\label{eq:app_marginal_mix}
\end{equation}
where $\Theta=\{\mu,\sigma^2,\pi\}$ denotes the collection of mixture means, variances, and weights.

\subsection{Variational Approximation and Objective}
\label{app:vi_objective}

DIVI applies a factorized variational approximation only to the global feature gates:
\begin{equation}
q(\boldsymbol{\phi})
=
\prod_{j=1}^D
\mathrm{Bernoulli}\!\big(\phi_j;\sigma(\eta_j)\big),
\label{eq:app_qphi}
\end{equation}
where $\eta_j$ are learnable variational logits and
\[
q_j \equiv q_\eta(\phi_j=1)=\sigma(\eta_j).
\]

Since the mixture parameters $\Theta=\{\mu,\sigma^2,\pi\}$ are optimized as point estimates while $\boldsymbol{\phi}$ is treated variationally, DIVI uses a hybrid variational objective rather than a full mean-field posterior over all latent quantities. Specifically, the implementation minimizes
\begin{equation}
\mathcal{J}
=
-\sum_{i=1}^N
\widehat{\mathbb{E}}_{q_T(\boldsymbol{\phi})}
\big[\log p(x_i\mid \Theta,\boldsymbol{\phi})\big]
+
\beta \sum_{j=1}^D
\mathrm{KL}\!\big(q(\phi_j)\,\|\,p(\phi_j)\big).
\label{eq:app_objective}
\end{equation}
This should be interpreted as a \emph{scaled variational objective} on the feature gates rather than as a joint ELBO over $(\mathbf Z,\boldsymbol{\phi})$.

The scaling $\beta=N$ is used in the implementation so that the KL penalty remains commensurate with the data log-likelihood, whose magnitude grows on the order of $N$. This reduces the tendency of the prior influence on feature gates to vanish as the sample size increases.

\paragraph{Analytical KL divergence}
Because $q(\boldsymbol{\phi})$ factorizes over dimensions, the KL term is available in closed form:
\begin{equation}
\mathrm{KL}\!\big(q(\phi_j)\,\|\,p(\phi_j)\big)
=
q_j \log\frac{q_j}{\rho_j}
+
(1-q_j)\log\frac{1-q_j}{1-\rho_j},
\label{eq:app_kl_bern}
\end{equation}
where $q_j=\sigma(\eta_j)$ and $p(\phi_j)=\mathrm{Bernoulli}(\rho_j)$. In code, both $q_j$ and $\rho_j$ are clamped away from $0$ and $1$ for numerical stability.

\paragraph{Monte Carlo estimator via Gumbel--Sigmoid relaxation}
Consistent with the implementation, the expectation term in Eq.~\eqref{eq:app_objective} is approximated using a one-sample Gumbel--Sigmoid relaxation related to the Gumbel--Softmax / Concrete reparameterization approaches \citep{jang2016categorical,maddison2016concrete}. Let $u_j \sim \mathrm{Uniform}(0,1)$ and define standard Gumbel noise
\[
g_j = -\log\!\big(-\log u_j\big).
\]
The relaxed gate is
\begin{equation}
\tilde{\phi}_j
=
\sigma\!\left(\frac{\eta_j+g_j}{T}\right),
\label{eq:app_gumbel_sample}
\end{equation}
where $T>0$ is the current temperature. The expectation term is then approximated by
\begin{equation}
\widehat{\mathbb{E}}_{q_T}
\big[\log p(x_i\mid \Theta,\boldsymbol{\phi})\big]
\approx
\log p(x_i\mid \Theta,\tilde{\boldsymbol{\phi}}).
\label{eq:app_mc_est}
\end{equation}

Given the relaxed sample $\tilde{\boldsymbol{\phi}}$, the marginal mixture likelihood becomes
\begin{equation}
\log p(x_i\mid \Theta,\tilde{\boldsymbol{\phi}})
=
\log\sum_{k=1}^K
\pi_k \exp\!\big(\ell_{ik}(\tilde{\boldsymbol{\phi}})\big),
\label{eq:app_mix_marginal}
\end{equation}
where
\begin{equation}
\ell_{ik}(\tilde{\boldsymbol{\phi}})
=
\sum_{j=1}^D
\Big[
\tilde{\phi}_j
\log \mathcal{N}(x_{ij}\mid \mu_{kj},\sigma_{kj}^2)
+
(1-\tilde{\phi}_j)
\log \mathcal{N}(x_{ij}\mid \mu_{0j},\sigma_{0j}^2)
\Big].
\label{eq:app_cluster_ll}
\end{equation}
This matches the implementation directly: the sampled mask is used to form weighted log-probabilities, after which a log-sum-exp is applied across mixture components.

Crucially, no explicit variational factor $q(\mathbf Z)$ is introduced. Cluster assignments remain implicit in the mixture likelihood, while the variational approximation is applied only to the global feature gates.

\subsection{Cluster Diagnostics and Split Criterion}
\label{app:split_rule}

DIVI grows structure monotonically through a split-only mechanism. Diagnostics are computed every $T_{\mathrm{split}}$ epochs.

\paragraph{Hard assignments for diagnostics}
Given current parameters and a deterministic relaxed gate vector, hard assignments are computed from the component-wise gated log-densities:
\begin{equation}
\hat z_i
=
\arg\max_k \ell_{ik}(\tilde{\boldsymbol{\phi}}),
\label{eq:app_hard_assign}
\end{equation}
where $\ell_{ik}$ is defined in Eq.~\eqref{eq:app_cluster_ll}. In the implementation, these assignments are formed using the deterministic gate probabilities $\sigma(\eta_j)$ rather than a stochastic sample.

\paragraph{Per-cluster average NLL score}
Let $\mathcal{I}_k=\{i:\hat z_i=k\}$. The diagnostic score for cluster $k$ is
\begin{equation}
S_k
=
-\frac{1}{|\mathcal{I}_k|}
\sum_{i\in\mathcal{I}_k}
\ell_{ik}(\tilde{\boldsymbol{\phi}}).
\label{eq:app_cluster_score}
\end{equation}
The worst-fitting cluster is
\[
k^\star = \arg\max_k S_k.
\]
If $S_{k^\star}>\tau$, a split is triggered.

\paragraph{Default threshold calibration}
When the threshold is not specified manually, the implementation uses the dimension-aware default
\begin{equation}
\tau
=
\frac{1}{2}D\Big(1+\log(2\pi)+\log \sigma^2\Big),
\label{eq:app_tau_default}
\end{equation}
with $\sigma^2=1$ by default. This corresponds to the entropy of a $D$-dimensional standard Gaussian up to the same scaling used in the implementation, and serves as a baseline calibration rather than a claim of universally optimal thresholding.

\paragraph{Split initialization}
When splitting cluster $k^\star$, the implementation constructs a new model with one additional component. The mean of the target cluster is duplicated and perturbed:
\[
\mu_a=\mu_{k^\star}+\delta_a,
\qquad
\mu_b=\mu_{k^\star}-\delta_b,
\qquad
\delta_a,\delta_b \sim \mathcal{N}(0,\sigma_{\text{split}}^2 I),
\]
where $\sigma_{\text{split}}$ is a small constant (default $0.2$). The corresponding log-variances are copied to both child components, and the feature-gating logits $\eta$ are inherited globally. The mixture mass of the split component is divided equally between the two new components, while all other components remain unchanged.

\subsection{Step A: Data-Informed Prior Construction}
\label{app:stepA}

Step~A constructs the prior inclusion probabilities $\rho_j$ used to initialize the global feature gates.

\paragraph{Mode 1: informative prior}
In the informative mode, Step~A first obtains a rough partition using K-means. For each feature $j$, it then computes:
\begin{enumerate}
    \item a rank-based between-group separation statistic using the Kruskal--Wallis test, and
    \item a Gaussian likelihood-ratio proxy that compares the rough cluster-wise fit against a pooled fit.
\end{enumerate}
The two scores are combined linearly, min-max normalized across features, and mapped through a logistic transform to obtain $\rho_j \in (0,1)$. In the implementation, the logistic contrast factor is approximately $6$, and the resulting probabilities are clamped away from $0$ and $1$.

\paragraph{Mode 2: noninformative prior}
In the noninformative mode, all features are initialized with
\[
\rho_j = 0.5.
\]

\paragraph{Mode 3: random prior}
In the random mode, each feature prior is initialized independently by
\[
\rho_j \sim \mathrm{Uniform}(0,1).
\]

This initialization stage is used only to shape the prior on the feature gates. All subsequent training is performed through the variational objective in Eq.~\eqref{eq:app_objective}.

\clearpage
\section{Additional Robustness and Sensitivity Results}
\label{app:supp_results}

\setcounter{equation}{0}
\setcounter{table}{0}
\setcounter{figure}{0}

\subsection{Detailed Sensitivity Tables}
\label{app:sensitivity_tables}

The following tables report detailed sensitivity summaries under fixed synthetic datasets. 
For compact presentation, results are separated into Panel A (large-$n$) and Panel B (small-$n$), 
and each row reports mean (standard deviation) over repeated runs.

\begin{table}[!htbp]
\centering
\footnotesize
\setlength{\tabcolsep}{3.5pt}
\renewcommand{\arraystretch}{0.96}
\caption{Sensitivity of DIVI to the split interval $T_{\mathrm{split}}$ on fixed synthetic datasets. Results are mean (standard deviation) over repeated runs.}
\label{tab:sens_tsplit}
\begin{tabular}{lcccccc}
\toprule
$T_{\mathrm{split}}$ & ARI & NMI & F1 & Time (s) & Final $K$ & Splits \\
\midrule
\multicolumn{7}{l}{\textit{Panel A: Large-$n$ ($N=1000$, $D=100$)}}\\
10 & 0.414 (0.096) & 0.618 (0.048) & 1.000 (0.000) & 15.21 (2.79) & 31.0 (0.0) & 30.0 (0.0) \\
20 & 0.493 (0.105) & 0.665 (0.045) & 1.000 (0.000) & 6.77 (0.23)  & 16.0 (0.0) & 15.0 (0.0) \\
40 & 0.678 (0.103) & 0.772 (0.039) & 1.000 (0.000) & 3.68 (0.25)  & 8.0  (0.0) & 7.0  (0.0) \\
80 & 0.879 (0.012) & 0.909 (0.006) & 1.000 (0.000) & 2.43 (0.31)  & 4.0  (0.0) & 3.0  (0.0) \\
\midrule
\multicolumn{7}{l}{\textit{Panel B: Small-$n$ ($N=200$, $D=100$)}}\\
10 & 0.187 (0.022) & 0.514 (0.014) & 0.593 (0.023) & 3.52 (0.51) & 31.0 (0.0) & 30.0 (0.0) \\
20 & 0.287 (0.031) & 0.583 (0.011) & 0.596 (0.028) & 1.89 (0.19) & 16.0 (0.0) & 15.0 (0.0) \\
40 & 0.507 (0.024) & 0.707 (0.011) & 0.572 (0.024) & 1.31 (0.14) & 8.0  (0.0) & 7.0  (0.0) \\
80 & 0.873 (0.002) & 0.907 (0.001) & 0.556 (0.013) & 0.87 (0.07) & 4.0  (0.0) & 3.0  (0.0) \\
\bottomrule
\end{tabular}
\end{table}

\begin{table}[!htbp]
\centering
\footnotesize
\setlength{\tabcolsep}{3.5pt}
\renewcommand{\arraystretch}{0.96}
\caption{Sensitivity of DIVI to the KL scaling factor $\beta$ on fixed synthetic datasets. Results are mean (standard deviation) over repeated runs.}
\label{tab:sens_beta}
\begin{tabular}{lcccccc}
\toprule
$\beta/N$ & ARI & NMI & F1 & Time (s) & Final $K$ & Active dims \\
\midrule
\multicolumn{7}{l}{\textit{Panel A: Large-$n$ ($N=1000$, $D=100$)}}\\
0.25 & 0.876 (0.008) & 0.908 (0.004) & 0.519 (0.017) & 2.46 (0.29) & 4.0 (0.0) & 28.6 (1.3) \\
0.50 & 0.877 (0.010) & 0.909 (0.005) & 0.828 (0.031) & 2.59 (0.27) & 4.0 (0.0) & 14.2 (0.9) \\
1.00 & 0.879 (0.012) & 0.909 (0.006) & 1.000 (0.000) & 2.53 (0.20) & 4.0 (0.0) & 10.0 (0.0) \\
2.00 & 0.880 (0.012) & 0.910 (0.006) & 1.000 (0.000) & 2.51 (0.24) & 4.0 (0.0) & 10.0 (0.0) \\
4.00 & 0.880 (0.012) & 0.910 (0.006) & 1.000 (0.000) & 2.53 (0.23) & 4.0 (0.0) & 10.0 (0.0) \\
\midrule
\multicolumn{7}{l}{\textit{Panel B: Small-$n$ ($N=200$, $D=100$)}}\\
0.25 & 0.868 (0.012) & 0.897 (0.018) & 0.245 (0.005) & 1.20 (0.43) & 4.0 (0.0) & 71.5 (1.6) \\
0.50 & 0.872 (0.008) & 0.903 (0.009) & 0.339 (0.007) & 1.03 (0.15) & 4.0 (0.0) & 49.0 (1.2) \\
1.00 & 0.873 (0.002) & 0.907 (0.001) & 0.556 (0.013) & 0.98 (0.09) & 4.0 (0.0) & 26.0 (0.8) \\
2.00 & 0.873 (0.002) & 0.907 (0.001) & 0.730 (0.014) & 0.88 (0.14) & 4.0 (0.0) & 17.4 (0.5) \\
4.00 & 0.873 (0.003) & 0.907 (0.002) & 0.778 (0.015) & 1.00 (0.17) & 4.0 (0.0) & 15.7 (0.5) \\
\bottomrule
\end{tabular}
\end{table}

\begin{table}[!htbp]
\centering
\footnotesize
\setlength{\tabcolsep}{3.2pt}
\renewcommand{\arraystretch}{0.92}
\caption{Sensitivity of DIVI to the entropy-threshold multiplier $\tau/\tau_0$ on fixed synthetic datasets. Results are mean (standard deviation) over repeated runs.}
\label{tab:sens_tau}
\begin{tabular}{lcccccc}
\toprule
$\tau/\tau_0$ & ARI & NMI & F1 & Time (s) & Final $K$ & Active dims \\
\midrule
\multicolumn{7}{l}{\textit{Panel A: Large-$n$ ($N=1000$, $D=100$)}}\\
0.9 & 0.879 (0.012) & 0.909 (0.006) & 1.000 (0.000) & 2.57 (0.20) & 4.0 (0.0) & 10.0 (0.0) \\
1.0 & 0.879 (0.012) & 0.909 (0.006) & 1.000 (0.000) & 2.54 (0.14) & 4.0 (0.0) & 10.0 (0.0) \\
1.1 & 0.879 (0.012) & 0.909 (0.006) & 1.000 (0.000) & 2.74 (0.23) & 4.0 (0.0) & 10.0 (0.0) \\
1.2 & 0.879 (0.012) & 0.909 (0.006) & 1.000 (0.000) & 2.67 (0.18) & 4.0 (0.0) & 10.0 (0.0) \\
\midrule
\multicolumn{7}{l}{\textit{Panel B: Small-$n$ ($N=200$, $D=100$)}}\\
0.9 & 0.873 (0.002) & 0.907 (0.001) & 0.556 (0.013) & 1.15 (0.56) & 4.0 (0.0) & 26.0 (0.8) \\
1.0 & 0.873 (0.002) & 0.907 (0.001) & 0.556 (0.013) & 0.99 (0.18) & 4.0 (0.0) & 26.0 (0.8) \\
1.1 & 0.873 (0.002) & 0.907 (0.001) & 0.556 (0.013) & 0.94 (0.15) & 4.0 (0.0) & 26.0 (0.8) \\
1.2 & 0.873 (0.002) & 0.907 (0.001) & 0.556 (0.013) & 0.99 (0.16) & 4.0 (0.0) & 26.0 (0.8) \\
\bottomrule
\end{tabular}
\end{table}

\begin{table}[!htbp]
\centering
\footnotesize
\setlength{\tabcolsep}{3.2pt}
\renewcommand{\arraystretch}{0.92}
\caption{Sensitivity of DIVI to the optimizer learning rate on fixed synthetic datasets. Results are mean (standard deviation) over repeated runs.}
\label{tab:sens_lr}
\begin{tabular}{lcccccc}
\toprule
Learning rate & ARI & NMI & F1 & Time (s) & Final $K$ & Active dims \\
\midrule
\multicolumn{7}{l}{\textit{Panel A: Large-$n$ ($N=1000$, $D=100$)}}\\
0.005 & 0.901 (0.027) & 0.915 (0.021) & 1.000 (0.000) & 2.34 (0.17) & 4.0 (0.0) & 10.0 (0.0) \\
0.010 & 0.879 (0.012) & 0.909 (0.006) & 1.000 (0.000) & 2.58 (0.26) & 4.0 (0.0) & 10.0 (0.0) \\
0.020 & 0.876 (0.008) & 0.908 (0.004) & 0.995 (0.015) & 2.80 (0.33) & 4.0 (0.0) & 10.1 (0.3) \\
0.050 & 0.877 (0.009) & 0.909 (0.004) & 0.990 (0.020) & 3.00 (0.21) & 4.0 (0.0) & 10.2 (0.4) \\
\midrule
\multicolumn{7}{l}{\textit{Panel B: Small-$n$ ($N=200$, $D=100$)}}\\
0.005 & 0.865 (0.019) & 0.888 (0.023) & 0.633 (0.014) & 0.98 (0.49) & 4.0 (0.0) & 21.6 (0.7) \\
0.010 & 0.873 (0.002) & 0.907 (0.001) & 0.556 (0.013) & 0.94 (0.17) & 4.0 (0.0) & 26.0 (0.8) \\
0.020 & 0.874 (0.003) & 0.907 (0.001) & 0.525 (0.017) & 0.95 (0.15) & 4.0 (0.0) & 28.1 (1.2) \\
0.050 & 0.875 (0.004) & 0.906 (0.004) & 0.505 (0.038) & 0.99 (0.13) & 4.0 (0.0) & 29.8 (2.8) \\
\bottomrule
\end{tabular}
\end{table}

\begin{table}[!htbp]
\centering
\footnotesize
\setlength{\tabcolsep}{3.5pt}
\renewcommand{\arraystretch}{0.92}
\caption{Sensitivity of DIVI to the temperature annealing endpoint $T_{\min}$ on fixed synthetic datasets. Results are mean (standard deviation) over repeated runs.}
\label{tab:sens_temp_end}
\begin{tabular}{lcccccc}
\toprule
$T_{\min}$ & ARI & NMI & F1 & Time (s) & Final $K$ & Active dims \\
\midrule
\multicolumn{7}{l}{\textit{Panel A: Large-$n$ ($N=1000$, $D=100$)}}\\
0.01 & 0.879 (0.012) & 0.910 (0.006) & 1.000 (0.000) & 2.80 (0.21) & 4.0 (0.0) & 10.0 (0.0) \\
0.05 & 0.879 (0.012) & 0.910 (0.006) & 1.000 (0.000) & 2.61 (0.23) & 4.0 (0.0) & 10.0 (0.0) \\
0.10 & 0.879 (0.012) & 0.909 (0.006) & 1.000 (0.000) & 2.55 (0.21) & 4.0 (0.0) & 10.0 (0.0) \\
0.20 & 0.880 (0.013) & 0.910 (0.007) & 1.000 (0.000) & 2.52 (0.14) & 4.0 (0.0) & 10.0 (0.0) \\
\midrule
\multicolumn{7}{l}{\textit{Panel B: Small-$n$ ($N=200$, $D=100$)}}\\
0.01 & 0.873 (0.003) & 0.907 (0.001) & 0.684 (0.028) & 1.16 (0.36) & 4.0 (0.0) & 19.3 (1.2) \\
0.05 & 0.873 (0.003) & 0.907 (0.001) & 0.572 (0.026) & 0.96 (0.15) & 4.0 (0.0) & 25.0 (1.5) \\
0.10 & 0.873 (0.002) & 0.907 (0.001) & 0.556 (0.013) & 1.07 (0.13) & 4.0 (0.0) & 26.0 (0.8) \\
0.20 & 0.873 (0.002) & 0.907 (0.001) & 0.568 (0.010) & 1.04 (0.15) & 4.0 (0.0) & 25.2 (0.6) \\
\bottomrule
\end{tabular}
\end{table}

\FloatBarrier
\clearpage
\subsection{Full Ablations for Misspecified Robustness Settings}
\label{app:misspec_full}

This subsection reports the full misspecified robustness results, including DIVI-Info, 
DIVI-NonInfo, and DIVI-Random, together with the oracle-$K$ K-means and oracle-$K$ GMM baselines. 
The purpose is to verify that the qualitative conclusions reported in the main text remain unchanged 
when the uninformed DIVI variants are shown explicitly.

\begin{table}[!htbp]
\centering
\footnotesize
\setlength{\tabcolsep}{4pt}
\renewcommand{\arraystretch}{0.95}
\caption{Full robustness results under heavy-tailed signal. All DIVI variants were evaluated with the sensitivity-selected split schedule $(\tau_{\mathrm{mult}}, T_{\mathrm{split}})=(1.00,120)$. Values are mean (standard deviation) over 20 independently generated datasets.}
\label{tab:misspec_full_heavytail}
\setlength{\tabcolsep}{3.5pt}
\begin{tabular}{llcccc}
\toprule
Method & $N$ & Final $K$ & ARI & NMI & F1 \\
\midrule
K-means (oracle $K$)   & 200  & 3.00 (0.00) & 0.913 (0.034) & 0.886 (0.036) & -- \\
Std.\ GMM (oracle $K$) & 200  & 3.00 (0.00) & 0.853 (0.238) & 0.875 (0.187) & -- \\
DIVI-Info              & 200  & 3.00 (0.00) & \textbf{0.989} (0.015) & \textbf{0.983} (0.022) & \textbf{0.768} (0.095) \\
DIVI-NonInfo           & 200  & 3.00 (0.00) & 0.906 (0.173) & 0.910 (0.124) & 0.182 (0.000) \\
DIVI-Random            & 200  & 3.00 (0.00) & 0.814 (0.190) & 0.819 (0.161) & 0.187 (0.041) \\
\midrule
K-means (oracle $K$)   & 1000 & 3.00 (0.00) & \textbf{0.986} (0.005) & \textbf{0.974} (0.009) & -- \\
Std.\ GMM (oracle $K$) & 1000 & 3.00 (0.00) & 0.971 (0.097) & 0.971 (0.066) & -- \\
DIVI-Info              & 1000 & 3.00 (0.00) & 0.964 (0.107) & 0.961 (0.080) & \textbf{0.990} (0.020) \\
DIVI-NonInfo           & 1000 & 3.00 (0.00) & 0.931 (0.168) & 0.941 (0.109) & 0.182 (0.000) \\
DIVI-Random            & 1000 & 3.00 (0.00) & 0.885 (0.193) & 0.890 (0.140) & 0.190 (0.041) \\
\bottomrule
\end{tabular}
\end{table}

\begin{table}[!htbp]
\centering
\footnotesize
\setlength{\tabcolsep}{4pt}
\renewcommand{\arraystretch}{0.95}
\caption{Full robustness results under correlated noise. All DIVI variants were evaluated with the sensitivity-selected split schedule $(\tau_{\mathrm{mult}}, T_{\mathrm{split}})=(1.00,120)$. Values are mean (standard deviation) over 20 independently generated datasets.}
\label{tab:misspec_full_corrnoise}
\setlength{\tabcolsep}{3.5pt}
\begin{tabular}{llcccc}
\toprule
Method & $N$ & Final $K$ & ARI & NMI & F1 \\
\midrule
K-means (oracle $K$)   & 200  & 3.00 (0.00) & 0.283 (0.080) & 0.307 (0.084) & -- \\
Std.\ GMM (oracle $K$) & 200  & 3.00 (0.00) & 0.307 (0.176) & 0.349 (0.194) & -- \\
DIVI-Info              & 200  & 3.00 (0.00) & \textbf{0.419} (0.153) & \textbf{0.466} (0.154) & \textbf{0.243} (0.038) \\
DIVI-NonInfo           & 200  & 3.00 (0.00) & 0.300 (0.170) & 0.341 (0.189) & 0.182 (0.000) \\
DIVI-Random            & 200  & 3.00 (0.00) & 0.195 (0.206) & 0.220 (0.217) & 0.181 (0.044) \\
\midrule
K-means (oracle $K$)   & 1000 & 3.00 (0.00) & 0.381 (0.037) & 0.407 (0.038) & -- \\
Std.\ GMM (oracle $K$) & 1000 & 3.00 (0.00) & 0.497 (0.007) & 0.565 (0.009) & -- \\
DIVI-Info              & 1000 & 3.00 (0.00) & \textbf{0.712} (0.261) & \textbf{0.731} (0.244) & \textbf{0.254} (0.031) \\
DIVI-NonInfo           & 1000 & 3.00 (0.00) & 0.464 (0.210) & 0.502 (0.210) & 0.182 (0.000) \\
DIVI-Random            & 1000 & 3.00 (0.00) & 0.297 (0.266) & 0.334 (0.276) & 0.182 (0.045) \\
\bottomrule
\end{tabular}
\end{table}

\end{document}